# Leaf Image based Plant Disease Identification using Color and Texture Features

Nisar Ahmed*, Hafiz Muhammad Shahzad Asif, Gulshan Saleem

***Abstract*—Identification of plant disease is usually done through visual inspection or during laboratory examination which causes delays resulting in yield loss by the time identification is complete. On the other hand, complex deep learning models perform the task with reasonable performance but due to their large size and high computational requirements, they are not suited to mobile and handheld devices. Our proposed approach contributes automated identification of plant diseases which follows a sequence of steps involving pre-processing, segmentation of diseased leaf area, calculation of features based on the Gray-Level Co-occurrence Matrix (GLCM), feature selection and classification. In this study, six color features and twenty-two texture features have been calculated. Support vector machines is used to perform one-vs-one classification of plant disease. The proposed model of disease identification provides an accuracy of 98.79% with a standard deviation of 0.57 on 10-fold cross validation. The accuracy on a self-collected dataset is 82.47% for disease identification and 91.40% for healthy and diseased classification. The reported performance measures are better or comparable to the existing approaches and highest among the feature-based methods, presenting it as the most suitable method to automated leaf-based plant disease identification. This prototype system can be extended by adding more disease categories or targeting specific crop or disease categories.**

## I. Introduction

Inconsistency and delay in the identification of plant diseases cause a reduction in the quantity and quality of yield. Losses due to plant diseases or other pest accounts for 20 to 40% of global annual productivity [1]. Studies have been carried out to assess the estimated loss caused by different diseases [1]. Yield loss also contributes toward increased consumer prices and a drop in the earnings for crop producers. Accurate and timely identification of plant diseases is crucial for ensuring maximum yield and is beneficial for farmlands in remote areas.

Advancements in machine vision have made it possible to perform the tasks of visual identification and these visual recognition methods can be employed for successful identification of plant diseases [2]. Image-based disease management and surveys have a long history of more than 90 years when aerial images were used to study crop disease [3]. Disease detection and identification have improved since then and informative and sophisticated analysis is being carried out [4]. Image based disease identification is under continuous development and recent

Nisar Ahmed, Department of Computer Engineering, University of Engineering and Technology Lahore, Pakistan. E-Mail: nisarahmedrana@yahoo.com Contact: +92-300-7272402 reviews can demonstrate this [5-7]. Moreover, studies have shown that image processing methods are effective to identify plant species or disease from leaf image [5-7]. Imaging devices have become cheaper and common with better quality images. Reliability, accuracy and precision of machine identification tasks have also continued to improve. As 60-70% of diseases appear exclusively on plant leaves which paves a way towards leaf image based plant disease identification [8].

This research is aimed at providing a solution for small farm holders in remote areas where expert knowledge is unavailable. The proposed work presents an approach to design a system for disease identification. The proposed work can be transformed into a mobile application which can perform the task of disease identification through visual classification. The major reason of crop yield loss is the attack of pests or various diseases so the risk of crop loss can be reduced by providing a solution of disease identification in farms [9]. The main objectives of the work are to identify whether a leaf is diseased or healthy, and classify the type of the disease to a diseased leaf image. The objective is fulfilled based on visual classification and can be accomplished in scenarios where the leaf disease is visually distinguishable. The previous research in this domain is intended to identify the diseases from few crops or diseases with a small sample size. Therefore, they provide a solution of limited utility as plants are present in great variety. The PlantVillage is a dataset with 54,309 leaf samples which provides 38 class labels for 14 crops [10]. This public dataset has a large training set with class labels which can be used by many researchers.

A feature-based approach is proposed here which tries to overcome the limitations of previous work in terms of accuracy and portability and the essence of the research lies in provision of high accuracy and portability to mobile devices. The proposed approach is a prototype system which is based on PlantVillage database having 38 classes. The system is able to identify the leaf as healthy and diseased and then categorize it into one of the diseases, present in the database. In order to identify additional symptoms or disease, such cases need to be added in the database and retraining the system will enable to identify those disease categories as well.

## II. Literature Review

Different approaches to identification and quantification of plant disease are in practice and leaf image-based identification of plant disease is one of them [11-17]. It is by far the easiest way to automatically identify plant disease and can be used for identification of various diseases [12]. The occurrence of plant disease causes specific changes in the texture and color of the leaf and therefore leaf imagery can be used to extract color and texture-based features to train a classifier. Some of the significant literature in the domain of plant leaf-based disease identification is provided below.

Nisar Ahmed, Department of Computer Engineering, University of Engineering and Technology Lahore, Pakistan. E-Mail: nisarahmedrana@yahoo.com Contact: +92-300-7272402

There are two approaches for leaf image based plant disease identification: (i) deep learning based, which use complex architectures to automatically learn features (ii) feature-based, which extract hand-crafted features such as color and texture features to train a conventional machine learning algorithm. The deep learning based approaches has provided higher accuracies but they require more computation and therefore not suitable for mobile or handheld devices with limited memory and computations. Some of the designed systems are targeted different diseases of some specific plant, whereas the other approaches target multiple plant diseases. Phadikar et al. [18] has presented a feature based approach to disease identification of rice plant. They have used Fermi energy based method for segmentation followed by color, contour and locality mapping. Rough set theory is used for selection of important features and rule mining with 10-fold cross-validation is used for system testing. Baquero et al. [19] has presented a Content-Based Image Retrieval (CBIR) system which uses color structure descriptors and nearest neighbors to classify important diseases or disease symptoms such as chlorosis, sooty molds and early blight. Similarly, Patil et al. [20] has also presented a CBIR and extracted color, shape and texture based features. Sandika et al. [21] has proposed a feature-based approach for disease identification of grapes leaves. They have also performed the comparison of texture feature's performance. of Their Oberti et al. [22] has targeted the fungal disease of grapevine plant (powdery mildew) due to its adverse effects on the crop yield and quality of produce. They have used multi-spectral imaging and captured grapevines leaf images at a range of angles (0 to 75 degrees). They have also highlighted that the detection sensitivity increases with the increase in angle and highest value is obtained at 60 degrees and for early middle ages the sensitivity improves from 9-75% with change in angle from 0-60 degrees. Similarly, Zhang et al. [15] has presented a feature-based approach which transform the image into superpixel representation and then segment the desired region using k-means and extract pyramid of histogram of orientation gradient (PHOG). Sharif et al. [23] has presented a feature-based approach for citrus fruit plant disease. They have used a hybrid feature selection technique based on principal component analysis and feature statistics. Singh et al. [24] has also presented a feature based approach for pine trees. Bai et al.[25] has targeted cucumber plant disease and proposed an improved fuzzy c-mean based clustering technique to segment the diseased leaf area. Hlaing et al. has presented a feature based approach and used PlantVillage dataset. The have extracted SIFT features and applied generalized pareto distributions to calculate density function. Support vector machines is used to train on these features and provided a 10-fold cross-validation accuracy of 84.7%.

Recently deep learning based approaches are used for leaf image based plant disease identification. Picon et al. [26] has used CNN for classification of plant diseases and claimed that deep learning based methods has provided highest performance. They have used Deep Residual Neural Networks and mainly targeted septoria, tan spot and

Nisar Ahmed, Department of Computer Engineering, University of Engineering and Technology Lahore, Pakistan. E-Mail: nisarahmedrana@yahoo.com Contact: +92-300-7272402 rust. Ferentinos et al. [12] has also presented a deep learning based solution using extended PlantVillage dataset with 58 classes. They have used pre-trained VGG for transfer learning and provided an accuracy of 99.53%. The cross-dataset evaluation has provided sharp decrease of 25-35% in accuracy indicating poor generalization. Mohanty et al. [14] and Yuan et al. [27] has proposed deep learning based solutions using PlantVillage dataset. They have used pre-trained CNN and applied transfer learning to classify plants into 38 classes. Zeng et al. [28] has presented a high-order residual CNN architecture which extracts low level details as well as high-level abstract representation simultaneously to improve classification performance and provided 91.3% classification accuracy with good generalization performance.

The proposed approach has targeted the problem through a two-step approach which is more suitable. PlantVillage dataset with 38 classes is used to demonstrate the proposed approach, and extended version of the dataset is also reported by some researchers but it was not available publicly at the time of this research. The first step is identification of the leaf as healthy or diseased. This step is performed with bag-of-features approach which is an effective method of visual classification. The further processing will only be performed on the diseased leaf to identify the type of disease affecting the plant leaf. This identification is performed by segmenting the diseased leaf region and extracted color and texture features. It is to be noted that we have extracted a comprehensive set of texture features which is not used in the literature for disease identification task. Feature normalization and selection is performed to obtain the most discriminatory feature set for classification. Five classification algorithms are tried in the final stage and Support Vector Machines (SVM) with cubic kernel is used as a final classifier. The proposed approach has provided comparable results to the state-of-the-art algorithms and demonstrate the effectiveness of texture based visual classification algorithms for the task of disease identification.

## III. Materials and Methods

Hughes et al. [8] has released a comprehensive dataset of plant diseases, expertly curated for leaf image-based plant disease identification. The dataset is named as PlantVillage and contains 54,309 images of diseased and healthy plant leaves and contain 38 class labels. These images cover 14 crop species: tomato, strawberry, squash, soybean, raspberry, potato, bell pepper, peach, orange, grape, corn, cherry and apple. These leaf images are affected by fungal, bacterial and viral disease. The dataset also contain images of 12 healthy species: apple, bell-pepper, blueberry, cherry, corn, grape, peach, potato, raspberry, soybean, strawberry and tomato. The leaf images are of 256×256 pixels' dimension with RGB colors.

Nisar Ahmed, Department of Computer Engineering, University of Engineering and Technology Lahore, Pakistan. E-Mail: nisarahmedrana@yahoo.com Contact: +92-300-7272402

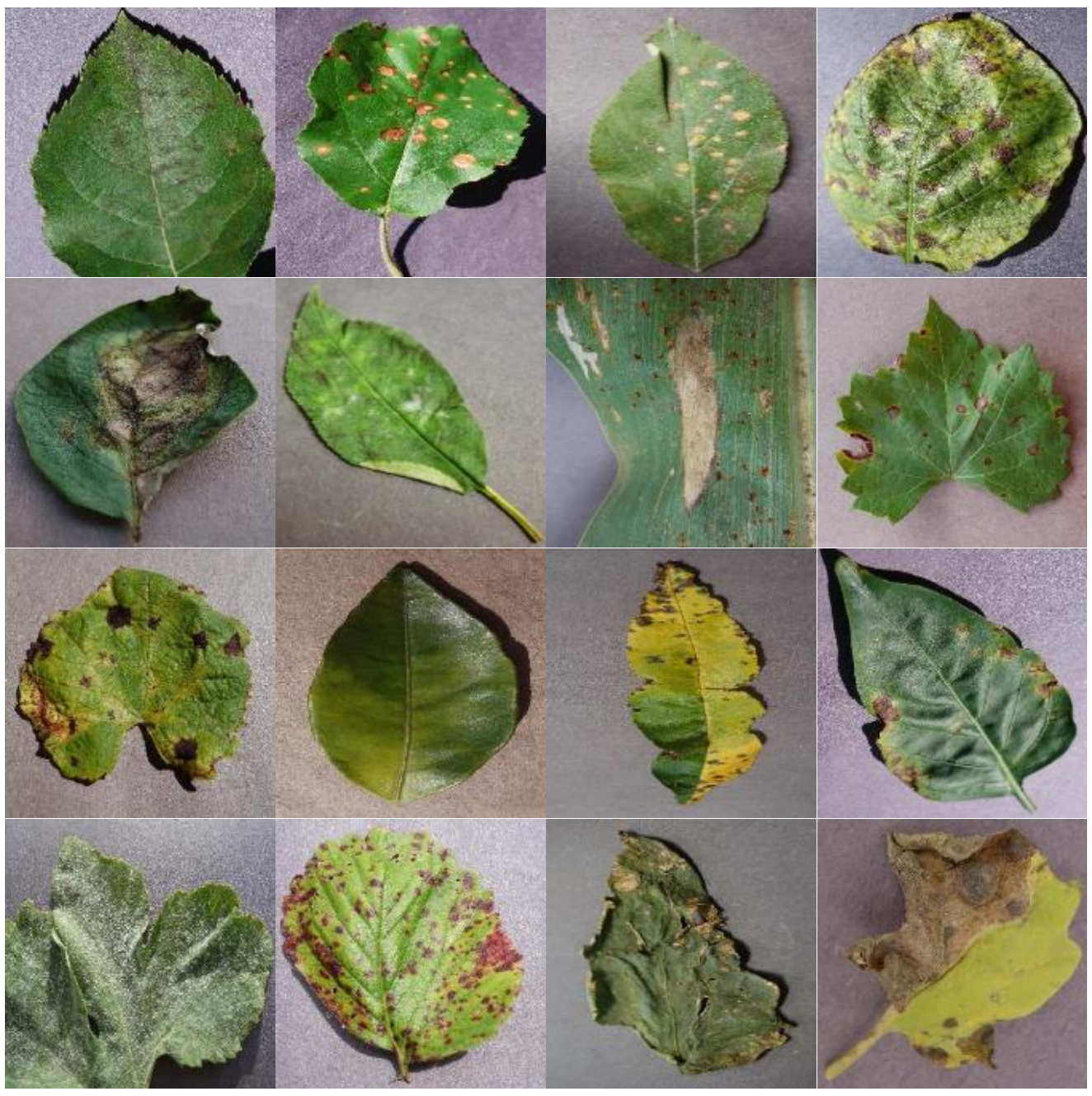

FIGURE 1 SOME RANDOMLY SELECTED LEAF IMAGES FROM PLANTVILLAGE DATASET

Figure 1 shows some sample leaf images from the dataset, these images are drawn randomly from the dataset in order to give a look at the raw form of images. These images are captured by trained staff in a regularized process and therefore contrast adjustment, color-cast and background removal is necessary to remove any possible bias. Table 1 highlights the 38 classes of PlantVillage dataset. There are 26 diseases occurring in 11 different plants and 12 healthy plant categories. The system would be able to identify any other disease or plant category if suitable number of training images are provided before final model construction. It is important to highlight that the proposed system will only be able to categories the image when the images are visually distinguishable as it follows a visual classification approach.

TABLE 1 PLANT AND THEIR CORRESPONDING DISEASE COVERED BY THE PLANTVILLAGE DATABASE

| | **Category** | **Plant** | **Disease** |
|---|---|---|---|
| 1 | | | Scab |
| 2 | | Apple | Black Rot |
| 3 | | | Cedar Apple Rust |
| 4 | | Bell Pepper | Bacterial Spot |
| 5 | | Cherry | Powdery Mildew |
| 6 | | | Cercospora Leaf Spot / Gray Leaf Spot |
| 7 | | Corn | Common Rust |
| 8 | | | Northern Leaf Blight |
| 9 | | | Black Rot |
| 10 | | Grape | Esca (Black Measles) |
| 11 | | | Leaf Blight (Isariopsis Leaf Spot) |
| 12 | Diseased | Orange | Haunglongbing (Citrus Greening) |
| 13 | | Peach | Bacterial Spot |
| 14 | | Potato | Early Blight |
| 15 | | | Late Blight |
| 16 | | Squash | Powdery Mildew |
| 17 | | Strawberry | Leaf Scorch |
| 18 | | | Bacterial Spot |
| 19 | | | Early Blight |
| 20 | | Tomato | Late Blight |
| 21 | | | Leaf Mold |
| 22 | | | Septoria Leaf Spot |

Nisar Ahmed, Department of Computer Engineering, University of Engineering and Technology Lahore, Pakistan. E-Mail: nisarahmedrana@yahoo.com Contact: +92-300-7272402

| | | | |
|---|---|---|---|
| 23 | | | Spider Mites (Two Spotted Spider Mites) |
| 24 | | | Target Spot |
| 25 | | | Mosaic Virus |
| 26 | | | Yellow Leaf Curl Virus |
| 27 | Healthy | Apple | No disease |
| 28 | | Bell Pepper | |
| 29 | | Blueberry | |
| 30 | | Cherry | |
| 31 | | Corn | |
| 32 | | Grape | |
| 33 | | Peach | |
| 34 | | Potato | |
| 35 | | Raspberry | |
| 36 | | Soybean | |
| 37 | | Strawberry | |
| 38 | | Tomato | |

The proposed approach is based on analysis of visual leaf features in a stepwise manner to construct a classifier and is provided below:

### A. *Background Removal*

The task of background removal is crucial in the approach as it may reduce the quality of features extracted from the leaf image. This step involves the removal of background to avoid any potential bias in extracted features and trained framework. Color cast removal is performed by normalizing the gray values of three-color channels separately. Moreover, the task of background removal needs to be automated and free from any human influence to increase the usability. The background removal can be performed in two ways: i) pixel clustering, and ii) edge detection. We have transformed RGB image to HSV (Hue, Saturation and Value) color space to easily separate background from leaf. Figure 2 provide the demonstration of true color image and its channels in RGB and HSV form to demonstrate that it is easier to transform the image to HSV and use Hue layer for background removal rather than performing the task of background removal on true color image itself.

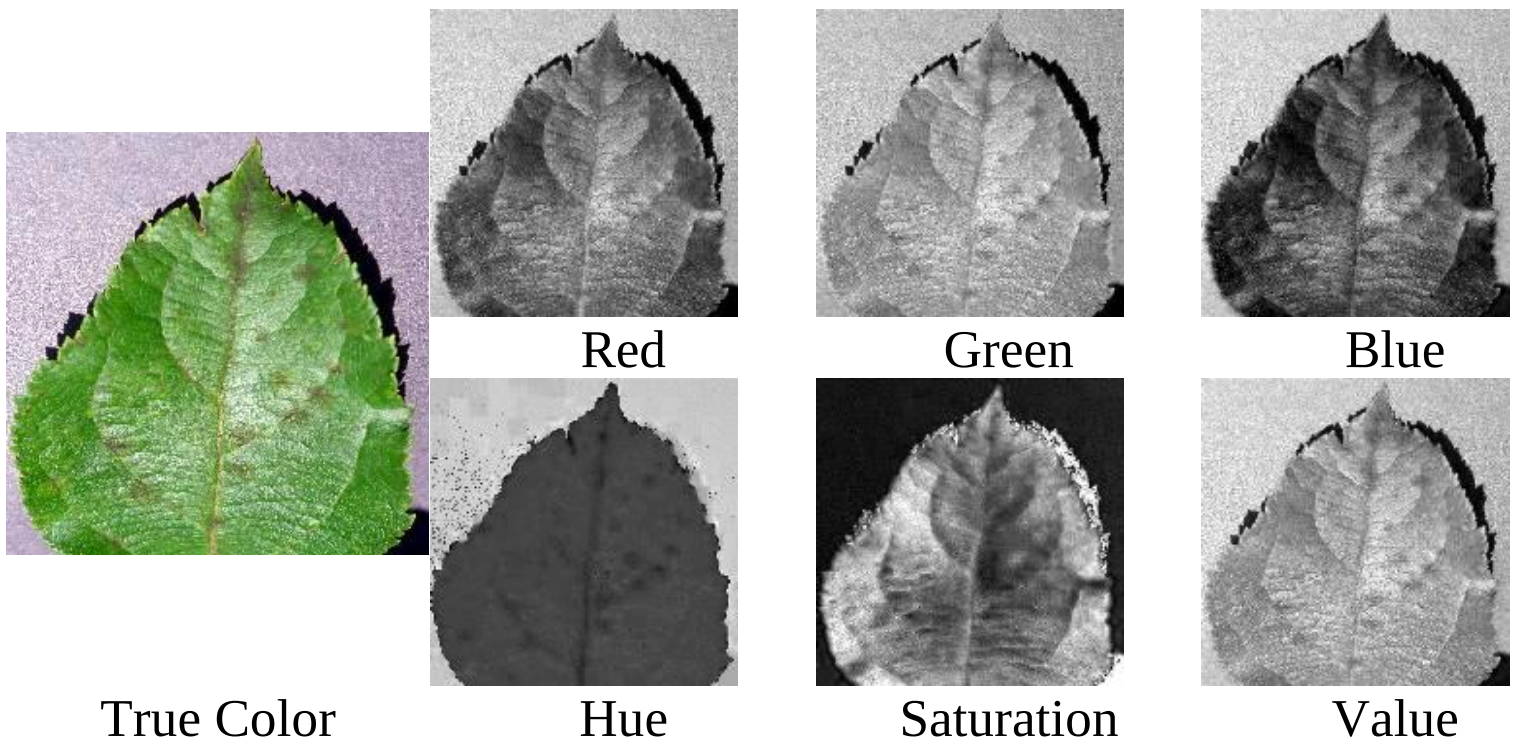

FIGURE 2 TRUE COLOR IMAGE (LEFT) RED, GREEN, BLUE AS WELL AS HUE, SATURATION AND VALUE CHANNELS OF THE IMAGE

The true color image and the color channels has shadows and light reflection which make them less suitable for background removal. This true color image is transformed to HSV color space and its three layers along with three RGB layers are displayed in Figure 2. It is clearly visible that Hue image is a better candidate for segmentation as it cast out all the intensity related information. Edge preserving blurring is applied on the Hue image using the

Nisar Ahmed, Department of Computer Engineering, University of Engineering and Technology Lahore, Pakistan. E-Mail: nisarahmedrana@yahoo.com Contact: +92-300-7272402 bilateral filtering and then Watershed algorithm is applied [29]. It finds watershed ridges in an image by treating light pixels as high elevations and dark pixels as low elevations. Eight neighborhood principle is used to segment adjoining regions. At the final stage, small pixel groups are removed by morphological processing which improved the mask. The stages of preprocessing and segmentation are provided in Figure 3.

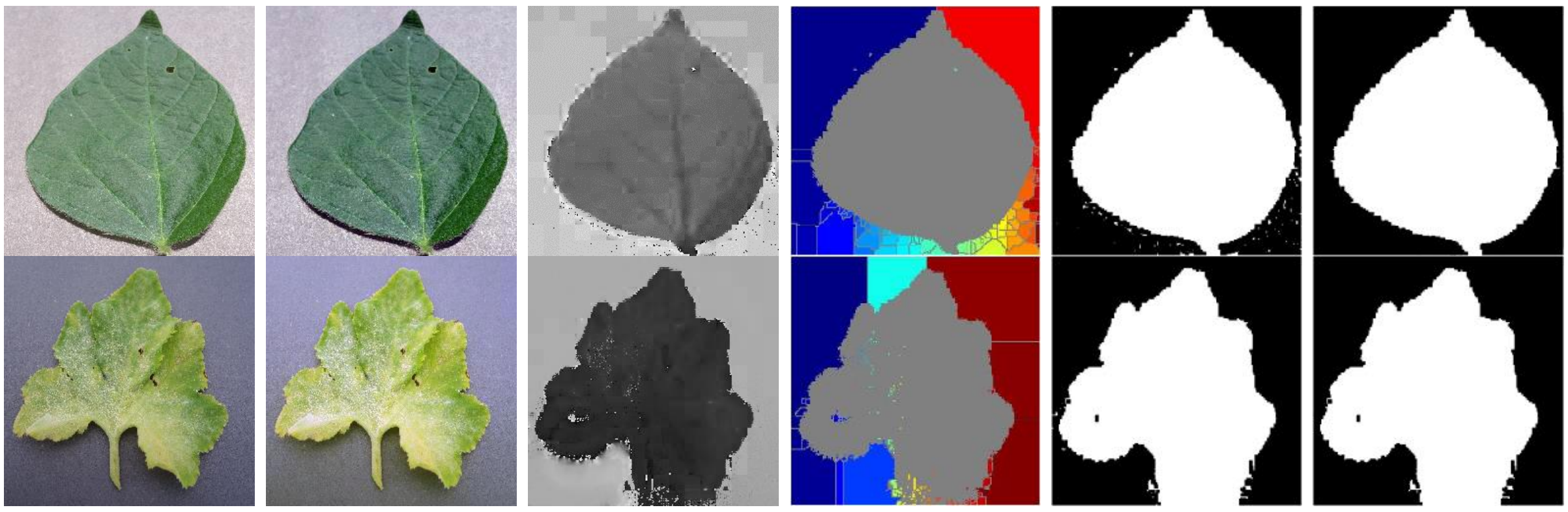

Unprocessed | Color Adjusted | Hue Layer | Watershed Transformed | Segmented Mask | Morphological Processed

*Figure 3 Steps of Preprocessing and background removal on two sample images*

Most of the images are perfectly segmented through this approach and few images had some extra area labelled as leaf. The segmentation results were slightly poor for Powdery Mildew of Squash shown lower in Figure 3. It is observed experimentally that slightly poor segmentation performance do not drastically affect the disease classification.

The background removal results of some sample leaf images are provided in Figure 4. It is important to note that most of the leaf images are perfectly segmented but few of the images are imperfectly segmented containing non-leaf area. Left most two images are the segmentation results of the leaf images of Figure 3. Lower-left image is of squash affected by powdery mildew and the images in this class are particularly poor for background removal.

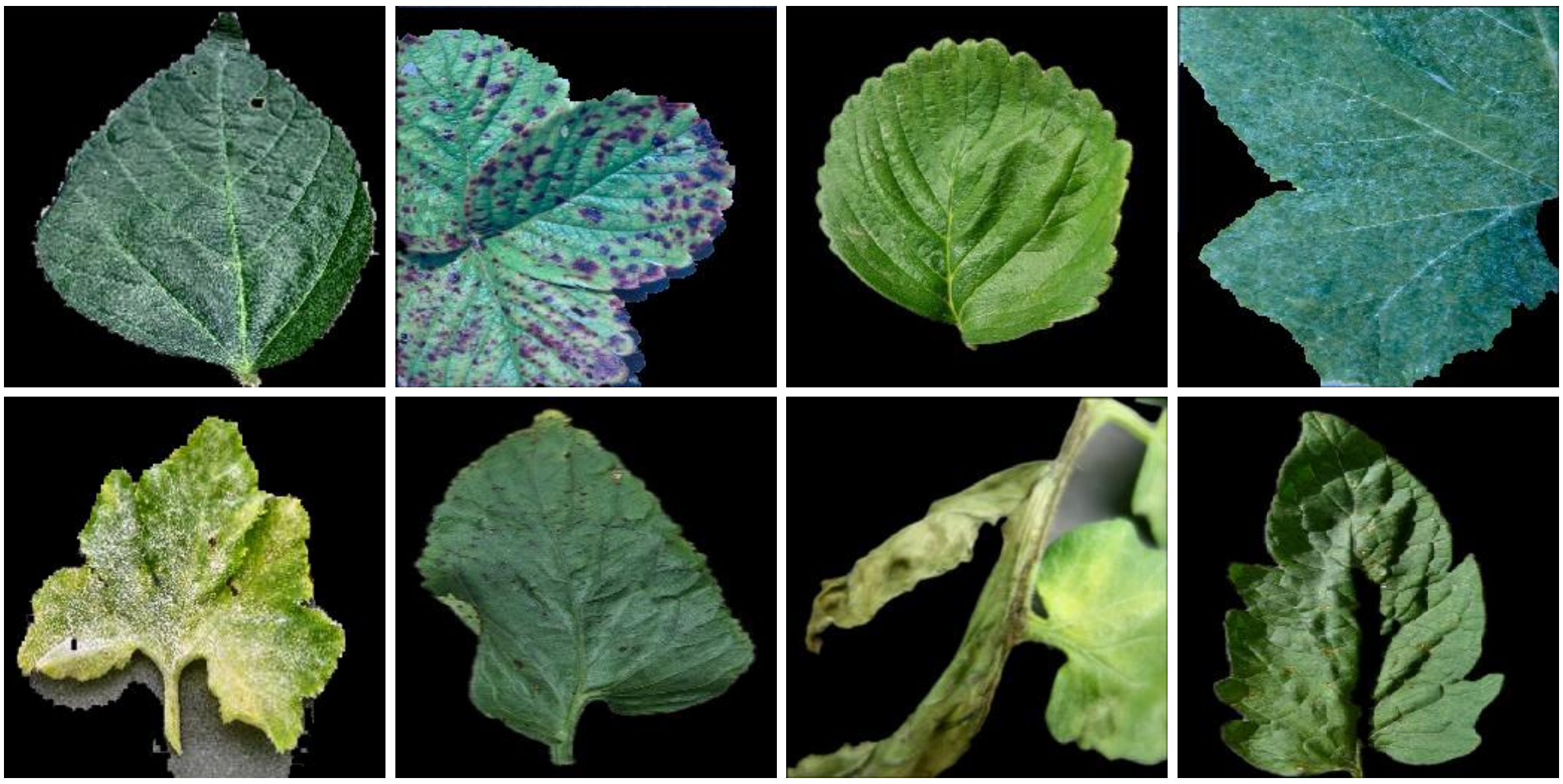

*Figure 4 Sample Images after background removal*

Nisar Ahmed, Department of Computer Engineering, University of Engineering and Technology Lahore, Pakistan. E-Mail: nisarahmedrana@yahoo.com Contact: +92-300-7272402

### B. *Healthy and Diseased Classification*

In PlantVillage dataset there were two categories of leaves: diseased and healthy. Disease identification involves clustering of the leaves in a number of regions, and segmentation of diseased image region. This segmented region has been processed to obtain texture and color-based features for disease identification. The healthy leaves do not have any diseased region and therefore, clustering and segmentation phase results into extraction of inconsistent leaf region and become the cause of degradation in identification of disease. The task of classification is separated into two steps: i) categorization into diseased and healthy leaves and ii) identification of plant disease among the 26 disease categories. The flow chart of this approach is shown in Figure 5.

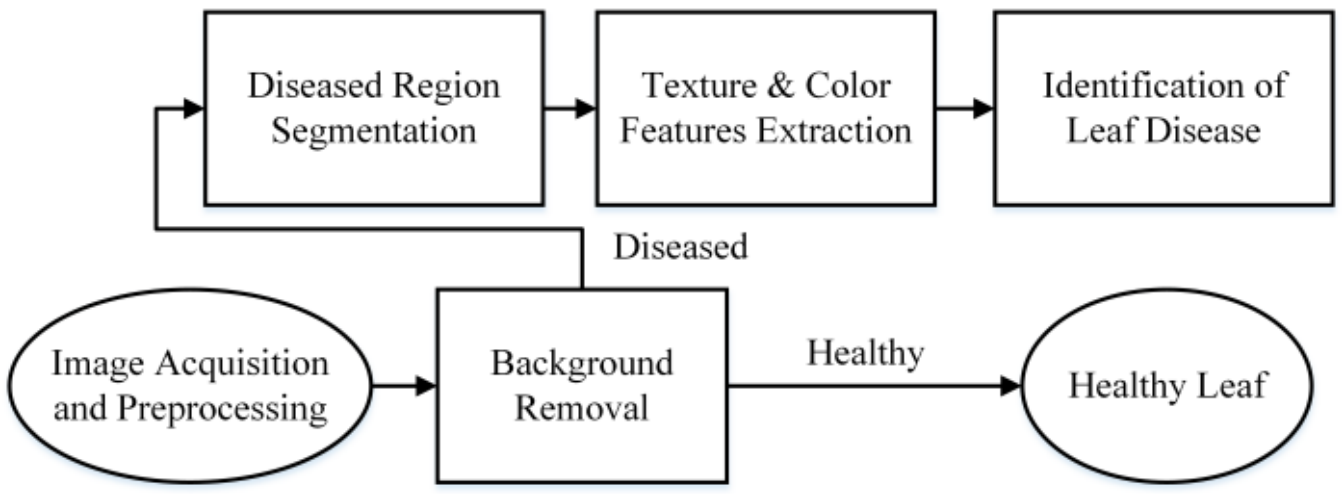

FIGURE 5 FLOW CHART OF LEAF DISEASE IDENTIFICATION

Classification of leaf image into healthy or diseased leaf is done by bag-of-features approach. Speeded Up Robust Features (SURF) are extracted from images which is a local feature detector and descriptor. First step of SURF detects interest point through blob detection. Blob are the region of image having points which are in some sense similar to each other and differ from its neighboring regions in term of color, contrast or brightness. These interest points can be of different scales and the feature descriptors provide a robust description of intensity distribution of the pixels within the neighborhood of the interest point. These features are used to construct a bag-of-features (inspired from bag of words approach of natural language processing) and a 50% of the images (27,155) are used for bag-of-features construction. The number of features are reduced through quantization of feature space using k-means clustering algorithm and the codebook size is kept small for portability to mobile devices. Moreover, 70% of the strongest features from each class (healthy or diseased) are used. The bag-of-features are used to train an SVM model having linear kernel with two classes to categorize the images as diseased or healthy.

### C. *Segmentation of Diseased Region*

Segmentation of diseased region is useful to extract discriminative features. In this study, diseased region is extracted for texture feature calculation as it contains the most discriminative portion of the plant disease. Therefore, extraction of diseased segments is a requirement which is described further. Among few options for diseased region segmentation we have opted Otsu's algorithm [30] which is computationally inexpensive and a reasonable choice overall. Otsu's algorithm is a simpler and quicker approach to segment the diseased region of

Nisar Ahmed, Department of Computer Engineering, University of Engineering and Technology Lahore, Pakistan. E-Mail: nisarahmedrana@yahoo.com Contact: +92-300-7272402 leaf. It works by calculating threshold in the gray-level histograms. It is assumed that the foreground and background pixels belong to two different Gaussian distributions with different mean and variance values. It finds the ridge between two peaks by maximizing the variance between the two classes and the final point in the gray-level histogram is used as a threshold to segment the image.

The diseased region segmentation is not exceptionally accurate however it is satisfactory. Figure 6 demonstrates the results of segmentation for Corn affected by common rust, Grape affected by leaf blight and potato affected by late blight.

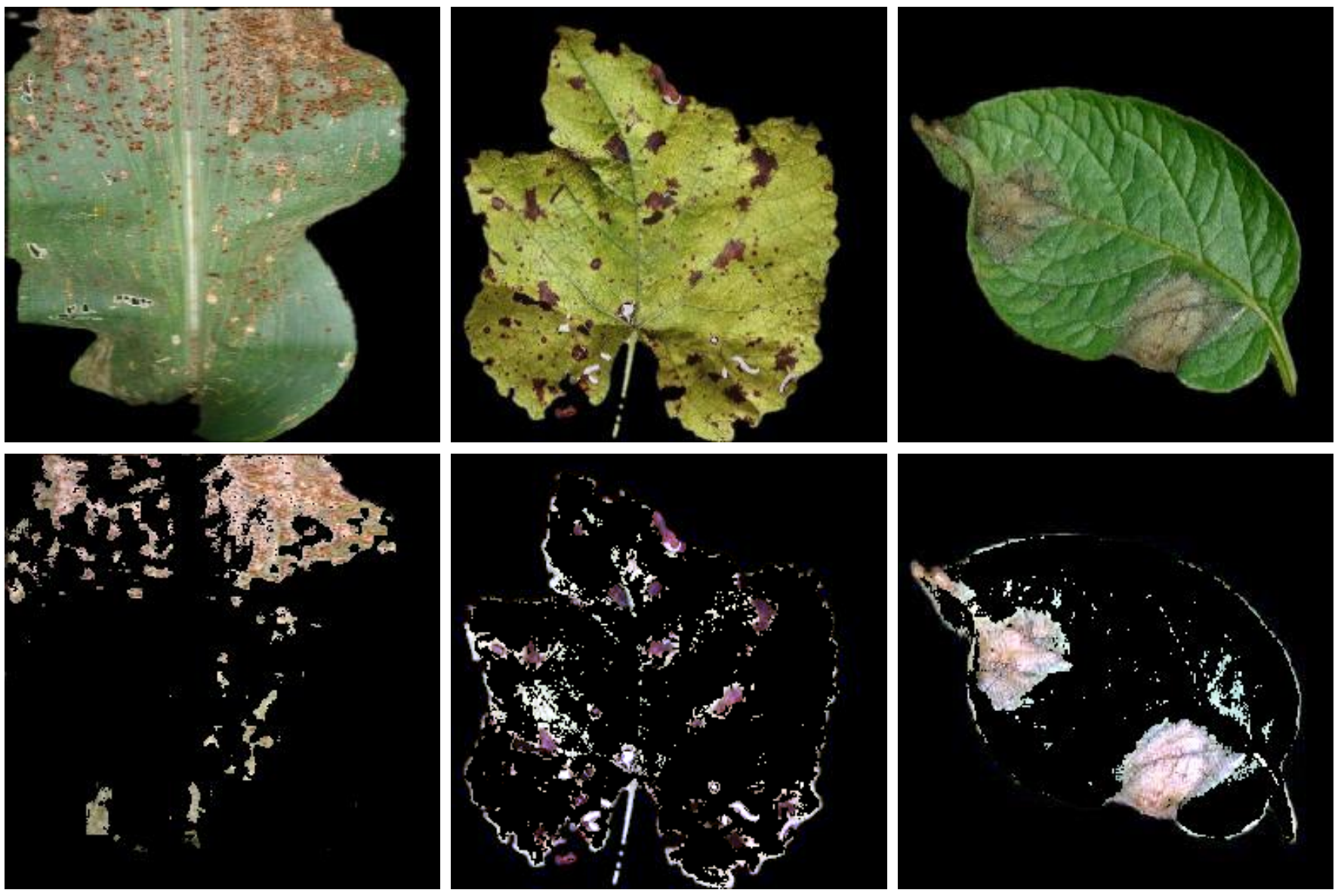

FIGURE 6 (ROW-1) DISEASED LEAF IMAGES OF CORN, GRAPE AND POTATO, (ROW-2) SEGMENTED DISEASED REGION OF LEAVES

### *D. Feature Extraction*

Feature extraction is an important step and the accuracy of a classification algorithm depends heavily on the feature set. Feature extraction is actually dimensionality reduction which is done to effectively represent the interesting parts of diseased region in a compact form. Shape, color and texture based features represent distinct leaf based classification features [31, 32]. The shape features can be used in identification of healthy or diseased leaves but in disease identification the segmented diseased region have different inter and intra-class variabilities and would be of less use. The color of the diseased region is distinct from healthy region and stays similar in intra-class samples and vary in inter-class samples. Similarly, the texture of the diseased region depends heavily on the type of disease and can be used as a major predictor. The parameters related to calculation of these two set of features are discussed below:

**Color Features**

There are two main types of color features namely: color histograms and color moments. In the present scenario, color is of lesser importance as predictor of disease categories. Color histograms provide a larger feature set which

Nisar Ahmed, Department of Computer Engineering, University of Engineering and Technology Lahore, Pakistan. E-Mail: nisarahmedrana@yahoo.com Contact: +92-300-7272402 is not greatly required in this scenario. Color moments are used to characterize color information which act as a compact way to represent color information.

I. Mean: Mean value of two-dimensional image matrix can be calculated to represent first color moments. The color image is separated into its three RGB layers and mean value is calculated for each layer.

II. Standard Deviation: Second color moment is standard deviation which represent the distribution of color information around the mean. It is calculated for three RGB layers for two-dimensional image matrix.

There is a total of six features representing color information in the diseased image region, three for mean and three for standard deviation of red, green and blue layers of RGB image.

**Texture Features**

Texture is more important part for disease identification as it represents more information related to the diseased region. Here 22 texture features have been extracted from Gray-Level Co-occurrence Matrix (GLCM) of grayscale leaf image as the color information is already encoded using color moments. The GLCM is calculated over grayscale image for which range is resampled into eight gray-levels. It calculates how frequently a pixel with a gray-level is situated adjacent to a pixel with the value $j$. The elements $(i, j)$ of the GLCM matrix of $8 \times 8$ represent the number of times that pixel with value $i$ occurred adjacent to a pixel with value $j$.

For an image with 8 different gray-level intensities, the GLCM $G$ of dimension $8 \times 8$ is defined over $m \times n$ image $I$, with an offset parameter $(\Delta_x, \Delta_y)$ formula in eq. 1 describes it.

$$G_{\Delta_x,\Delta_y}(i,j) = \sum_{x=1}^{m}\sum_{y=1}^{n}\begin{cases}1, & if\ I(x,y) = i\ and\ I(x+\Delta_x, y+\Delta_y) = j\\ 0, & otherwise\end{cases} \quad (1)$$

Where: $i$ & $j$ are the pixel values; $x$ and $y$ are the spatial positions in the image matrix $I$; $(\Delta_x, \Delta_y)$ are the offset which defines the spatial relation of the matrix and $I(x, y)$ indicates the pixel values at location $(x, y)$.

The details of these texture features is provided as follows:

| Sr. No. | Feature | Formula |
|---|---|---|
| 1 | Uniformity (angular second moment) [33-36] | $Uniformity = \sum_{i=1}^{N_g}\sum_{j=1}^{N_g} p(i,j)^2$ |
| 2 | Entropy [33-36] | $Entropy = -\sum_{i=1}^{N_g}\sum_{j=1}^{N_g} p(i,j) log[p(i,j)]$ |
| 3 | Contrast [33-36] | $Contrast = \sum_{i=1}^{N_g}(i-j)^2 \left\{\sum_{i=1}^{N_g}\sum_{j=1}^{N_g} p(i,j)\right\}$ |

Nisar Ahmed, Department of Computer Engineering, University of Engineering and Technology Lahore, Pakistan. E-Mail: nisarahmedrana@yahoo.com Contact: +92-300-7272402

| | | |
|---|---|---|
| 4 | Dissimilarity [33, 34, 36] | $Dissimilarity = \sum_{i=1}^{N_g}(i-j)\left\{\sum_{i=1}^{N_g}\sum_{j=1}^{N_g}p(i,j)\right\}$ |
| 5 | Inverse Difference Moment (Homogeneity) [33, 36] | $Homogeneity = \sum_{i=1}^{N_g}\sum_{j=1}^{N_g}\frac{p(i,j)}{1+(i-j)^2}$ |
| 6 | Inverse Difference (ID) [33] | $ID = \sum_{i=1}^{N_g}\sum_{j=1}^{N_g}\frac{p(i,j)}{1+\|i-j\|}$ |
| 7 | Correlation [36] | $Correlation = \sum_{i=1}^{N_g}\sum_{j=1}^{N_g}\frac{ijp(i,j)-\mu_x\mu_y}{\sigma_x\sigma_y}$ |
| 8 | Autocorrelation [34] | $Autocorrelation = \sum_{i=1}^{N_g}\sum_{j=1}^{N_g}p(i,j)\bar{p}(i-k,j-k)$<br>Where k is the amount of shift |
| 9 | Cluster Shade (CS) [33, 34] | $CS = \sum_{i=1}^{N_g}\sum_{j=1}^{N_g}p\left(i+j-\mu_x-\mu_y\right)^3\bar{p}(i,j)$ |
| 10 | Cluster Prominence (CP) [33, 34] | $\mathrm{CP} = \sum_{i=1}^{N_g}\sum_{j=1}^{N_g}p\left(i+j-\mu_x-\mu_y\right)^4\bar{p}(i,j)$ |
| 11 | Maximum Probability (MP) [33, 34] | $MP = \max_{i,j} p(i,j)$ |
| 12 | Sum of Squares (SS) [35] | $SS = \sum_{i=1}^{N_g}\sum_{j=1}^{N_g}p(i-\mu)^2p(i,j)$ |
| 13 | Sum Average (SA) [35, 36] | $SA = \sum_{i=2}^{2N_g} ip_{x+y}(i)$ |
| 14 | Sum Variance (SV) [35, 36] | $SV = \sum_{i=2}^{2N_g}\left(i-\left[\sum_{i=2}^{2N_g}p_{x+y}(i)\right]\right)^2$ |
| 15 | Sum Entropy (SE) [35, 36] | $SE = -\sum_{i=2}^{2N_g}p_{x+y}(i)log\{p_{x+y}(i)\}$ |
| 16 | Difference Variance (DV) [35, 36] | $\boldsymbol{DV} = \sum_{i=2}^{2N_g}\left(i-\left[\sum_{i=2}^{2N_g}ip_{x-y}(i)\right]\right)^2$ |
| 17 | Difference Entropy (DE) [35, 36] | $\boldsymbol{DE} = -\sum_{i=0}^{N_g-1}p_{x-y}(i)log\{p_{x-y}(i)\}$ |
| 18 | Information Measures of Correlation (IMC) [35] | $IMC_1 = \frac{HXY-HXY_1}{max\{HX,HY\}}$<br>$IMC_2 = \sqrt{1-e^{[-2.0(HXY_2-HXY)]}}$<br>Where; $HXY = \sum_{i=1}^{N_g}\sum_{j=1}^{N_g}p(i,j)\log\left(p(i,j)\right)$ |
| 19 | Maximal Correlation Coefficient (MCC) [35] | $\boldsymbol{MCC} = \sqrt{F\left(\sum_k\frac{p(i,k)p(j,k)}{p_x(i)p_y(k)}\right)}$<br>Where; $F$ is the second largest Eigen value |
| 20 | Inverse Difference Normalized (IDN) [33] | $\boldsymbol{IDN} = \sum_{i,j=1}^{G}\frac{C_{ij}}{1+\|i-j\|^2/G^2}$ |

Nisar Ahmed, Department of Computer Engineering, University of Engineering and Technology Lahore, Pakistan. E-Mail: nisarahmedrana@yahoo.com Contact: +92-300-7272402

| | | |
|---|---|---|
| 21 | Inverse Difference Moment Normalized (IDMN) [33] | $IDMN = \sum_{i,j=1}^{G} \frac{C_{ij}}{1 + (i-j)^2 / G^2}$ |

There is a total of 22 texture features which represent the texture information of the segmented diseased region of leaf image.

*E. Feature Scaling*

It is a method to scale the range of values of a feature vector. It is also known as feature normalization or standardization. Feature values are computed in different units or may represent different parameters so their ranges vary widely. It is to note that many classifiers use some distance measure, such as Euclidian distance, between the points. If one feature has wide range of values, the distance measure will be largely affected by this specific feature, so, the range of all the features should be normalized to reflect proportionality in the distance measure. Moreover, the convergence time of gradient descent is less with feature scaling and it is sometimes less for stochastic gradient descent as well. In SVM, feature scaling can reduce the time to find support vectors and affect the results as well. The performance of K-NN, SVM, logistic regression, perceptron, artificial neural networks and linear discriminant analysis is largely affected by feature scaling.

Different feature scaling methods are used in the literature and selection of a particular method depends on the requirements. Feature scaling usually is applied to change the range of features to a constrained value such as [0,255] in case of images or [0, 1] or [-1, 1] in case of features with real values. Max-min normalization is most widely used method in which features are scaled to a specified range i.e. [0, 1] by scaling their maximum and minimum values to these points. In this work, standardization is opted which scale the features such that they have zero mean and unit variance. This is widely used scaling method for SVM, ANN and linear regression [37]. Formula in eq. 2 is used for standardization of features:

$$x' = \frac{x - \bar{x}}{\sigma} \quad (2)$$

*F. Feature Selection*

Feature selection is employed to select a subset of features for model construction. There are four major reasons to perform feature selection: i) simplification of model, ii) shorter training time, iii) avoidance from curse of dimensionality [38] and (iv) reduction of over fitting. The basic idea behind feature selection is that feature set can contain irrelevant as well as redundant features. Redundant and irrelevant are different terms, as one relevant feature may be redundant in the presence of some other correlated feature.

Nisar Ahmed, Department of Computer Engineering, University of Engineering and Technology Lahore, Pakistan. E-Mail: nisarahmedrana@yahoo.com Contact: +92-300-7272402

Texture and color features have lot of dependencies on each other and have high correlation in some cases. Inclusion of all the features may cause degradation in performance and will result in slower training. Therefore, feature selection is crucial. There are different approaches for feature selection and we have explored two of them for our problem: i) sequential feature selection and ii) ReliefF algorithm. Sequential feature selection can be used in two configurations, one is forward and the other is backward. The Forward Feature Selection (FFS) starts from one feature and iteratively adds features and compares the performance based on predefined criteria. The features are added till the accuracy stop increasing or start decreasing. In backward feature elimination, the process is started with all the features and one feature is removed at a time and performance is compared. This work has used forward feature selection in which the training algorithm is trained and cross-validated. We have used SVM with 10-fold Cross-Validation (CV) for evaluation and the feature with maximum accuracy are retained. The process of feature inclusion is continued till there is increase in accuracy. Then 19 features have been selected in the given order and achieved a CV accuracy of 0.934 at the end, Table 2 provides the list of these features and the CV accuracy.

TABLE 2: FEATURE SELECTED THROUGH FORWARD FEATURE SELECTION WITH MULTI-SVM AS CRITERION

| Sr. No. | Feature Name | Cross-Validation Accuracy |
|---|---|---|
| 1 | Correlation | 0.704019 |
| *2* | Autocorrelation | 0.718989 |
| *3* | Sum of Squares | 0.728080 |
| *4* | Standard Deviation of Blue | 0.760980 |
| *5* | Maximal correlation coefficient | 0.768141 |
| 6 | Information measures of correlation (2) | 0.770524 |
| 7 | Mean of Blue | 0.775902 |
| *8* | Sum Variance | 0.783225 |
| *9* | Sum Entropy | 0.802779 |
| *10* | Cluster Prominence | 0.802957 |
| *11* | Standard Deviation of Red | 0.821492 |
| 12 | Uniformity | 0.831278 |
| 13 | Contrast | 0.851472 |
| 14 | Difference entropy | 0.859829 |
| *15* | Homogeneity | 0.880204 |
| *16* | Entropy | 0.908171 |
| *17* | Mean of Green | 0.914150 |
| *18* | Inverse difference | 0.930732 |
| *19* | Inverse difference moment normalized | 0.934014 |

ReliefF algorithm is a feature ranking algorithm and this ranking can be truncated based on weights to select a subset of features. It finds weights of the features in the cases where target dependent variable is a multiclass categorical variable. The algorithm rewards the features that give different values to neighbors of different classes and penalize the features that give different value to neighbors of same class. In initialization all the feature weights are set to zero and then a random observation is iteratively selected. K-nearest neighbor's algorithm is used to select the nearest observations to randomly selected observation and updates the weights for all the classes for that feature. The algorithm provides two measures as an output, feature ranks and weights. The ranking is an integer array with index number of each feature ordered in decreasing rank. The weights array contains the prediction weights for all the ranked vectors. Negative weights indicate that the feature may do more harm than good in classification performance. Eight features out of 29 have positive weight values and rest of the 20 features have negative or zero weight values. We have discarded the features with negative weights and selected the ones having positive weight values. These eight selected features along with their rank and weight are provided in the Table 3.

The two feature subsets, one with eight features and the other with nineteen features were used to check their classification performance and it was observed that the features selected with FFS is better suited for classifier training.

TABLE 3: SELECTED FEATURES ALONG WITH THEIR RANK AND WEIGHTS

| Rank | Feature Name | Feature Weight |
|---|---|---|
| 1 | Standard Deviation of Blue | 0.0015 |
| *2* | Mean of Blue | 0.0010 |
| *3* | Standard Deviation of Red | 0.0008 |
| *4* | Standard Deviation of Green | 0.0007 |
| *5* | Mean of Red | 0.0007 |
| 6 | Inverse difference | 0.0002 |
| 7 | Entropy | 0.0002 |
| *8* | Difference entropy | 0.0002 |

### *G. Classifier Selection*

The final stage of the work is selection of suitable classification algorithm for classification of leaf disease to the category they belong. We have chosen five major classification algorithms to check their suitability for classification. The algorithm with best performance is optimized for its hyper-parameters to form a final model. This study has used multi-SVM, K-NN, Naïve Bayes, Random Forest and Artificial Neural Networks (ANN).

Multi-SVM: The SVM is a supervised learning algorithm that is used for classification, regression, and clustering or outlier detection. Basically, SVM is a binary linear classifier which separate two classes using maximum margin hyperplane. A good separation is provided by a hyperplane which has largest distance to the nearest data points, therefore referred as maximum margin hyperplane, as larger the margin lower the generalization error. The feature set may exist in a finite dimensional space but it is not linearly separable in that space. To make the data linearly separable it is transformed to a higher dimensional space using a kernel function where it is linearly separable.

The binary classification nature of SVM is overcome by several methods one-vs-all and one-vs-one are two of them. In one-vs-all multi-SVM, all the classes are selected separately and one is taken as positive label and all other classes as negative labels and a total of N classifiers are trained. The problem with imbalanced data is that one class have less number of samples then all the other classes and general SVM implementation may behave poorly. One-vs-one train separate classifier for each class paired with all others, one by one, and thus overcoming the problem of imbalance. This leads to $\frac{N(N-1)}{2}$ classifiers making it much more computationally expensive.

Nisar Ahmed, Department of Computer Engineering, University of Engineering and Technology Lahore, Pakistan. E-Mail: nisarahmedrana@yahoo.com Contact: +92-300-7272402

The present study has used the one-vs-one multi-SVM and tried linear, quadratic, cubic and Gaussian kernels whereas cubic SVM provided best cross-validation performance. The kernel functions used for SVM are provided in Table 4.

K-NN: It is a non-parametric algorithm which is used for classification, regression or outlier detection. It is a lazy learning algorithm which doesn't try to construct a model, rather simply stores the training examples of the data. The classification is performed through simple majority voting of the k-nearest neighbors. This is a preferred algorithm for noisy or large training data and is easier to implement. The problem with K-NN is selection of the value of K. Smaller value of K makes finer decision boundary resulting in overfitting, and larger K value results in smoother boundary resulting in poor classification accuracy due to higher bias. Determination of suitable value of K is computationally expensive as it needs to compute distance of each example with all training samples.

TABLE 4: KERNEL FUNCTIONS USED WITH SVM

| Sr. No. | Kernel Method | Kernel Function |
|---|---|---|
| 1 | Linear | $K(x,y) = sum(x * y)$ |
| *2* | Polynomial | $K(x,y) = 1 + sum(x * y)^d$ |
| *3* | Gaussian | $K(x,y) = e^{-\left(\frac{\|x-y\|^2}{2\sigma^2}\right)}$ |

Naïve Bayes: It is a probabilistic classification algorithm based on Bayes theory having an assumption that each feature is independent from others. It has an apparently wrong and simplistic assumption that each feature is independent from the presence of any other feature in the feature set. This assumption also helps to alleviate the problems arising due to curse of dimensionality. However, Naïve Bayes have good performance in real-world situations, it require small amount of training data and is fast to compute.

Naïve Bayes is not a single algorithm but a set of algorithms for classification which are based on common assumption that a specific feature is independent of any other feature. It can be trained efficiently using supervised learning approach and in many setting parameter estimations are based on maximum likelihood.

Random Forest: Overfitting is the inherent problem of decision tree which is overcame using different bagging or boasting approaches. The problem of overfitting is addressed by random forest which fits a number of decision tree on several sub-samples of training dataset and use an average for improvement of the predictive accuracy. The sub-sample size is always same as the original sample size except the samples are drawn with replacement. The principle advantage of reduction of over-fitting make random forest more accurate than simple decision trees in most cases. However, it is a complex algorithm and its prediction speed is slow.

Artificial neural networks: ANNs are computational algorithms which are roughly inspired by the biological artificial neural networks. These systems learn to perform tasks from labelled examples, without any task specific rules. For a classification task, they will automatically adjust the weight of their neurons based on training

Nisar Ahmed, Department of Computer Engineering, University of Engineering and Technology Lahore, Pakistan. E-Mail: nisarahmedrana@yahoo.com Contact: +92-300-7272402 examples and try to classify new instances based on learned weights. A multi-layer artificial neural network has an input layer, an output layer and one or more hidden layers. Normally, the number of neurons in the input layer are equal to feature vector length and number of neurons in the output layer are equal to the number of classes. The number of hidden layers and the number of neurons in each hidden layer depends on the particular problem set and can't be computed beforehand.

In this study, we have 29 neurons in input layers, 26 neurons in the output layer and number of neurons in hidden layer are selected through search.

## IV. Results and Discussions

Existing studies have used feature-based or CNN-based classification to categorize the leaves of PlantVillage dataset into one of 38 categories. The proposed approach mainly relied on texture features extracted through statistics of GLCM applied on segmented diseased region of the leaf. As the segmentation stage of diseased leaf area is automated, when applied on healthy leaves it extracts some leaf area and confuses the classifier. The confusion matrix of the diseased or healthy plant classification stage is provided in Table 5:

TABLE 5 CONFUSION MATRIX OF HEALTHY AND DISEASED PLANT CLASSES

| | Healthy | Diseased |
|---|---|---|
| Healthy | 99.7848 | 0.2152 |
| Diseased | 0.8280 | 99.1720 |

Precision, recall, accuracy and F1-Score of the healthy and diseased plant classification are provided in Table 5. The results of the classification are reasonable but existing approaches have not used classification between healthy and diseased leaves for this dataset.

TABLE 6 EVALUATION RESULTS OF DISEASED AND HEALTHY LEAF CLASSIFICATION

| | Precision | Recall | Accuracy | F1-Score |
|---|---|---|---|---|
| SVM | 99.78 | 99.18 | 99.48 | 99.48 |

The second phase involves the segmentation of diseased leaf region of 39,226 images. Six color and twenty-two texture features were extracted and a feature vector with a length of 29 features is formed. This feature vector is standardized so that the feature vector has zero mean and unit variance. Standardization is applied to remove the feature range effect on distance-based classifiers, reduce model complexity and improve the gradient calculation. Feature selection using ReliefF algorithm and forward feature selection using Multi-SVM are used for feature subset selection. Here, eight features have been selected with ReliefF algorithm having positive weight values and nineteen features using FFS which provided improvement in accuracy. The evaluation of three feature sets, i) complete feature set, ii) sub-set with ReliefF algorithm and iii) sub-set with FFS is performed on five classifiers discussed in section IV (G).

Nisar Ahmed, Department of Computer Engineering, University of Engineering and Technology Lahore, Pakistan. E-Mail: nisarahmedrana@yahoo.com Contact: +92-300-7272402

It has been observed that the classification accuracy of subset selected using ReliefF algorithm is significantly less than the classification accuracy of the complete feature set. Whereas, sub-set with FFS provided slightly better classification accuracy than complete feature set. It has been a favorable decision which provides better performance in terms of time and classification accuracy. The results of classification evaluation with three feature sets are provided in Table 7.

TABLE 7 CLASSIFIER EVALUATION RESULTS WITH COMPLETE FEATURE SET AND TWO SUBSETS

| | Complete 29 features | ReliefF 8 features | FFS 19 features |
|---|---|---|---|
| Multi-SVM | 92.1 | 80.2 | 93.4 |
| K-NN | 84.5 | 85.5 | 86.7 |
| Naïve Bayes | 78.4 | 76.1 | 82.4 |
| Random Forest | 82.2 | 77.2 | 87.9 |
| ANN | 91.7 | 87.4 | 92.6 |

The feature subset selected after feature selection stage is used for fine tuning the Multi-SVM and ANN as they both provided highest accuracy in initial stage.

The ANN was configured with one hidden layer having neurons varying from 10 to 100 neurons and two hidden layers with neurons varying from 10 to 50 neurons and evaluated on the basis of Cross-Entropy (CE). The network with least cross-entropy has been tested for classification accuracy of which turned out to be 93.4% with the neural net of Figure 7.

TABLE 8 ARTIFICIAL NEURAL NETWORK PERFORMANCE WITH DIFFERENT NUMBER OF NEURONS

| Network Size One Hidden Layer | CE | Network Size Two Hidden Layers | CE |
|---|---|---|---|
| 10 | 0.0476 | 10:10 | 0.0489 |
| 20 | 0.0378 | 15:10 | 0.0454 |
| 30 | 0.0342 | 20:10 | 0.0429 |
| 40 | 0.0332 | 20:15 | 0.0376 |
| 50 | 0.0328 | 30:20 | 0.0352 |
| 60 | 0.0351 | 40:30 | 0.0340 |
| 70 | 0.0336 | 40:40 | 0.0325 |
| 80 | 0.0335 | 50:30 | 0.0332 |
| 90 | 0.0326 | 50:40 | 0.0327 |
| 100 | 0.0383 | 50:50 | 0.0313 |

Table 8 provides the network performance with different number of neurons. In neural net with one hidden layer, minimum cross-entropy of 0.0328 is achieved with 50 neurons in the hidden layer. For second neural net with two hidden layers, minimum cross-entropy of 0.0313 is achieved with 50 neurons in the first hidden layer and 50 neurons in the second hidden layer.

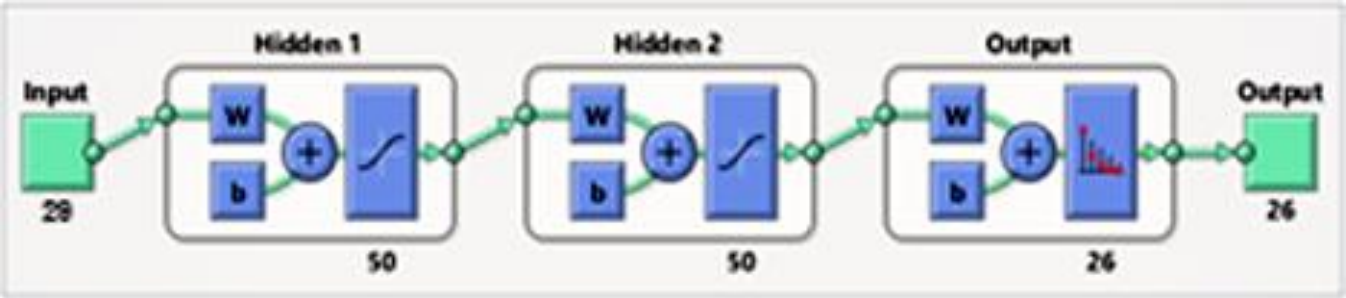

FIGURE 7 ARTIFICIAL NEURAL NETWORK ARCHITECTURE WITH TWO HIDDEN LAYERS

The multi-SVM with one-vs-one approach was used and cross-validation accuracy was compared for linear, quadratic, cubic and Gaussian kernels. The best CV accuracy is provided by multi-SVM with cubic kernel having

Nisar Ahmed, Department of Computer Engineering, University of Engineering and Technology Lahore, Pakistan. E-Mail: nisarahmedrana@yahoo.com Contact: +92-300-7272402 performance parameters provided as in Table 9. Note that the accuracy provided by SVM with cubic kernel is higher than the accuracy provided by the artificial neural network with 50 neurons in the first hidden layer and 50 neurons in the second hidden layer.

TABLE 9 PERFORMANCE PARAMETERS FOR SVM WITH DIFFERENT KERNELS, THE TIME IS TAKEN IN SECONDS AND PREDICTION SPEED IS TAKEN IN OBSERVATIONS PER SECONDS

| | Linear | Quadratic | Cubic | Gaussian |
|---|---|---|---|---|
| Accuracy | 84.39% | 95.19% | 98.99% | 83.89% |
| Training Time | 137s | 645s | 3177s | 107s |
| Prediction Speed | 3800 obs/s | 1300 obs/s | 1800 obs/s | 510 obs/s |

It can be observed that SVM has provided impressive performance with cubic kernel but it has taken longer training time (almost 53 minutes). The prediction speed for SVM with cubic kernel, however, is only slower than linear kernel. Since training is a one-time job and can be done on a faster machine, prediction speed is the desired parameter for its practical use. Therefore, SVM with cubic kernel was selected as the final model. The confusion matrix for the final model is provided in the Table 10 below:

Nisar Ahmed, Department of Computer Engineering, University of Engineering and Technology Lahore, Pakistan. E-Mail: nisarahmedrana@yahoo.com Contact: +92-300-7272402

TABLE 10 DISEASE CLASSIFICATION CONFUSION MATRIX

| | | | | | | | | | | | | | | | | | | | | | | | | | | |
|---|---|---|---|---|---|---|---|---|---|---|---|---|---|---|---|---|---|---|---|---|---|---|---|---|---|---|
| Apple_scab | 1.00 | 0.00 | 0.00 | 0.00 | 0.00 | 0.00 | 0.00 | 0.00 | 0.00 | 0.00 | 0.00 | 0.00 | 0.00 | 0.00 | 0.00 | 0.00 | 0.00 | 0.00 | 0.00 | 0.00 | 0.00 | 0.00 | 0.00 | 0.00 | 0.00 | 0.00 |
| Apple_Black_rot | 0.00 | 1.00 | 0.00 | 0.00 | 0.00 | 0.00 | 0.00 | 0.00 | 0.00 | 0.00 | 0.00 | 0.00 | 0.00 | 0.00 | 0.00 | 0.00 | 0.00 | 0.00 | 0.00 | 0.00 | 0.00 | 0.00 | 0.00 | 0.00 | 0.00 | 0.00 |
| Cedar_apple_rust | 0.00 | 0.00 | 1.00 | 0.00 | 0.00 | 0.00 | 0.00 | 0.00 | 0.00 | 0.00 | 0.00 | 0.00 | 0.00 | 0.00 | 0.00 | 0.00 | 0.00 | 0.00 | 0.00 | 0.00 | 0.00 | 0.00 | 0.00 | 0.00 | 0.00 | 0.00 |
| Cherry_Powdery_mildew | 0.00 | 0.00 | 0.00 | 0.99 | 0.00 | 0.00 | 0.00 | 0.00 | 0.00 | 0.00 | 0.00 | 0.00 | 0.00 | 0.00 | 0.01 | 0.00 | 0.00 | 0.00 | 0.00 | 0.00 | 0.00 | 0.00 | 0.00 | 0.00 | 0.00 | 0.00 |
| Gray_leaf_spot | 0.00 | 0.00 | 0.00 | 0.00 | 0.99 | 0.00 | 0.01 | 0.00 | 0.00 | 0.00 | 0.00 | 0.00 | 0.00 | 0.00 | 0.00 | 0.00 | 0.00 | 0.00 | 0.00 | 0.00 | 0.00 | 0.00 | 0.00 | 0.00 | 0.00 | 0.00 |
| Maize_Common_rust | 0.00 | 0.00 | 0.00 | 0.00 | 0.00 | 0.99 | 0.00 | 0.00 | 0.00 | 0.00 | 0.00 | 0.00 | 0.00 | 0.00 | 0.01 | 0.00 | 0.00 | 0.00 | 0.00 | 0.00 | 0.00 | 0.00 | 0.00 | 0.00 | 0.00 | 0.00 |
| Northern_Leaf_Blight | 0.00 | 0.00 | 0.00 | 0.00 | 0.00 | 0.00 | 0.94 | 0.00 | 0.00 | 0.00 | 0.00 | 0.00 | 0.00 | 0.00 | 0.05 | 0.00 | 0.00 | 0.00 | 0.00 | 0.00 | 0.00 | 0.00 | 0.00 | 0.00 | 0.00 | 0.00 |
| Grape_Black_rot | 0.00 | 0.00 | 0.00 | 0.00 | 0.00 | 0.00 | 0.00 | 1.00 | 0.00 | 0.00 | 0.00 | 0.00 | 0.00 | 0.00 | 0.00 | 0.00 | 0.00 | 0.00 | 0.00 | 0.00 | 0.00 | 0.00 | 0.00 | 0.00 | 0.00 | 0.00 |
| Grape_Black_Measles | 0.00 | 0.00 | 0.00 | 0.00 | 0.00 | 0.00 | 0.00 | 0.00 | 1.00 | 0.00 | 0.00 | 0.00 | 0.00 | 0.00 | 0.00 | 0.00 | 0.00 | 0.00 | 0.00 | 0.00 | 0.00 | 0.00 | 0.00 | 0.00 | 0.00 | 0.00 |
| Grape_Leaf_blight | 0.00 | 0.00 | 0.00 | 0.00 | 0.00 | 0.00 | 0.00 | 0.00 | 0.00 | 0.98 | 0.00 | 0.00 | 0.00 | 0.00 | 0.01 | 0.00 | 0.00 | 0.00 | 0.00 | 0.00 | 0.00 | 0.00 | 0.00 | 0.00 | 0.00 | 0.00 |
| Citrus_Greening | 0.00 | 0.00 | 0.00 | 0.00 | 0.00 | 0.00 | 0.00 | 0.00 | 0.00 | 0.00 | 0.97 | 0.00 | 0.00 | 0.00 | 0.03 | 0.00 | 0.00 | 0.00 | 0.00 | 0.00 | 0.00 | 0.00 | 0.00 | 0.00 | 0.00 | 0.00 |
| Peach_Bacterial_spot | 0.00 | 0.00 | 0.00 | 0.00 | 0.00 | 0.00 | 0.00 | 0.00 | 0.00 | 0.00 | 0.00 | 0.99 | 0.00 | 0.00 | 0.00 | 0.00 | 0.00 | 0.00 | 0.00 | 0.00 | 0.00 | 0.00 | 0.00 | 0.00 | 0.00 | 0.00 |
| Pepper_Bacterial_spot | 0.00 | 0.00 | 0.00 | 0.00 | 0.00 | 0.00 | 0.00 | 0.00 | 0.00 | 0.00 | 0.00 | 0.00 | 1.00 | 0.00 | 0.00 | 0.00 | 0.00 | 0.00 | 0.00 | 0.00 | 0.00 | 0.00 | 0.00 | 0.00 | 0.00 | 0.00 |
| Potato_Early_blight | 0.00 | 0.00 | 0.00 | 0.00 | 0.00 | 0.00 | 0.00 | 0.00 | 0.00 | 0.00 | 0.00 | 0.00 | 0.00 | 1.00 | 0.00 | 0.00 | 0.00 | 0.00 | 0.00 | 0.00 | 0.00 | 0.00 | 0.00 | 0.00 | 0.00 | 0.00 |
| Potato_Late_blight | 0.00 | 0.00 | 0.00 | 0.00 | 0.00 | 0.00 | 0.00 | 0.00 | 0.00 | 0.00 | 0.00 | 0.00 | 0.00 | 0.00 | 1.00 | 0.00 | 0.00 | 0.00 | 0.00 | 0.00 | 0.00 | 0.00 | 0.00 | 0.00 | 0.00 | 0.00 |
| Squash_Powdery_mildew | 0.00 | 0.00 | 0.00 | 0.00 | 0.00 | 0.00 | 0.00 | 0.00 | 0.00 | 0.00 | 0.00 | 0.00 | 0.00 | 0.00 | 0.02 | 0.98 | 0.00 | 0.00 | 0.00 | 0.00 | 0.00 | 0.00 | 0.00 | 0.00 | 0.00 | 0.00 |
| Strawberry_Leaf_scorch | 0.00 | 0.00 | 0.00 | 0.00 | 0.00 | 0.00 | 0.00 | 0.00 | 0.00 | 0.00 | 0.00 | 0.00 | 0.00 | 0.00 | 0.00 | 0.00 | 1.00 | 0.00 | 0.00 | 0.00 | 0.00 | 0.00 | 0.00 | 0.00 | 0.00 | 0.00 |
| Tomato_Bacterial_spot | 0.00 | 0.00 | 0.00 | 0.00 | 0.00 | 0.00 | 0.00 | 0.00 | 0.00 | 0.00 | 0.00 | 0.00 | 0.00 | 0.00 | 0.00 | 0.00 | 0.00 | 1.00 | 0.00 | 0.00 | 0.00 | 0.00 | 0.00 | 0.00 | 0.00 | 0.00 |
| Tomato_Early_blight | 0.00 | 0.00 | 0.00 | 0.00 | 0.00 | 0.00 | 0.00 | 0.00 | 0.00 | 0.00 | 0.00 | 0.00 | 0.00 | 0.00 | 0.00 | 0.00 | 0.00 | 0.00 | 1.00 | 0.00 | 0.00 | 0.00 | 0.00 | 0.00 | 0.00 | 0.00 |
| Tomato_healthy | 0.00 | 0.00 | 0.00 | 0.00 | 0.00 | 0.00 | 0.00 | 0.00 | 0.00 | 0.00 | 0.00 | 0.00 | 0.00 | 0.00 | 0.00 | 0.00 | 0.00 | 0.00 | 0.00 | 1.00 | 0.00 | 0.00 | 0.00 | 0.00 | 0.00 | 0.00 |
| Tomato_Late_blight | 0.00 | 0.00 | 0.00 | 0.00 | 0.00 | 0.00 | 0.00 | 0.00 | 0.00 | 0.00 | 0.00 | 0.00 | 0.00 | 0.00 | 0.00 | 0.00 | 0.00 | 0.00 | 0.00 | 0.00 | 1.00 | 0.00 | 0.00 | 0.00 | 0.00 | 0.00 |
| Tomato_Leaf_Mold | 0.00 | 0.00 | 0.00 | 0.00 | 0.00 | 0.00 | 0.00 | 0.00 | 0.00 | 0.00 | 0.00 | 0.00 | 0.00 | 0.00 | 0.00 | 0.00 | 0.00 | 0.00 | 0.00 | 0.00 | 0.00 | 1.00 | 0.00 | 0.00 | 0.00 | 0.00 |
| Septoria_leaf_spot | 0.00 | 0.00 | 0.00 | 0.00 | 0.00 | 0.00 | 0.00 | 0.00 | 0.00 | 0.00 | 0.00 | 0.00 | 0.00 | 0.00 | 0.00 | 0.00 | 0.00 | 0.00 | 0.00 | 0.00 | 0.00 | 0.00 | 0.99 | 0.00 | 0.00 | 0.00 |
| Tomato_Spider_mites | 0.00 | 0.00 | 0.00 | 0.00 | 0.00 | 0.00 | 0.00 | 0.00 | 0.00 | 0.00 | 0.00 | 0.00 | 0.00 | 0.00 | 0.00 | 0.00 | 0.00 | 0.00 | 0.00 | 0.00 | 0.00 | 0.00 | 0.00 | 1.00 | 0.00 | 0.00 |
| Tomato_Target_Spot | 0.00 | 0.00 | 0.00 | 0.00 | 0.00 | 0.00 | 0.00 | 0.00 | 0.00 | 0.00 | 0.00 | 0.00 | 0.00 | 0.00 | 0.00 | 0.00 | 0.00 | 0.00 | 0.00 | 0.00 | 0.00 | 0.00 | 0.00 | 0.00 | 1.00 | 0.00 |
| Tomato_mosaic_virus | 0.00 | 0.00 | 0.00 | 0.00 | 0.00 | 0.00 | 0.00 | 0.00 | 0.00 | 0.00 | 0.00 | 0.00 | 0.00 | 0.00 | 0.00 | 0.00 | 0.00 | 0.00 | 0.00 | 0.00 | 0.00 | 0.00 | 0.00 | 0.00 | 0.00 | 1.00 |

Nisar Ahmed, Department of Computer Engineering, University of Engineering and Technology Lahore, Pakistan. E-Mail: nisarahmedrana@yahoo.com Contact: +92-300-7272402

TABLE 11 PLANT DISEASE CLASSIFICATION EVALUATION RESULTS FOR DISEASE CLASSIFICATION

| | Precision | Recall | Accuracy | F1-Score |
|---|---|---|---|---|
| Proposed | 99.35 | 99.31 | 99.31 | 99.33 |

Table 11 provides the results of classification evaluation for the task of plant disease identification. Please note that precision is macro-average precision which is averaged across 26 classes. The recall is also average of the recall for each class. The proposed model has provided an accuracy of 99.31 and F1-score of 99.33 which is quite high and comparable to existing approaches. The high accuracy can be explained based on rich set of texture and color features and with the use of feature selection which provides subset of most discriminatory features.

It is to be noted that the results of Table 10 are for 26 diseased classes only whereas the complete dataset contains 38 classes. In the literature, only the leaf category is classified so the results of Table 10 can't be compared with them. However, we have calculated the aggregate classification performance by accumulating the classification scores of the both stages which is provided in Table 12. It is to be noted that final classification score is slightly less than the disease classification module performance provided in Table 11.

TABLE 12 PLANT DISEASE CLASSIFICATION EVALUATION FOR FINAL MODEL

| | Precision | Recall | Accuracy | F1-Score |
|---|---|---|---|---|
| Proposed | 99.13 | 98.49 | 98.79 | 98.81 |

The difference in classifier performance is due to their limitations such Naïve Bayes performs poor due to lack of all posterior probabilities, random forest can't perform good due to limited performance of decision tree and their averaging over the number of trees can give intermediate accuracy. K-NN can perform well in cases where features can be discriminated based on some distance metric and artificial neural network has provided comparable accuracy to SVM and the slightly lower accuracy is due to difficulty in finding the global minima. The artificial neural network has been trained numerous times and there is slight change in its accuracy which is due to the initialization of its weight and difficulty to find a global minimum. The ANN is iterated several times with different number of neurons in one and two hidden layers. The iterations are limited due to computational complexity and can be repeated more times if a good solution is not achieved using other classifiers.

Table 13 provides a comparison of the proposed method with existing work. Most of the methods has used train/test split for training and validation of their method whereas some methods have employed Cross-Validation (CV) in which the data is divided into $\frac{1}{n}$ folds and during each iteration $n-1$ folds are used for training and one of them is used for testing. After completion of n-folds the average values are obtained.

TABLE 13 COMPARISON WITH EXISTING METHODS

| Year | Existing Work | Accuracy | Precision | Recall | F1-Score | Remarks |
|---|---|---|---|---|---|---|
| 2016 | Mohanty et al. [14] | **99.35** | **99.35** | **99.35** | **99.34** | 38 classes with 60/40 split |
| 2017 | Islam et al. [39] | 95 | 95 | 95 | 95 | 3 classes with 60/40 split |
| 2017 | Hlaing et al. [40] | 78.7 | - | - | - | 38 classes with 5-fold CV |
| 2017 | Hlaing Et al. [41] | 84.7 | - | - | - | 7 classes with 10-fold CV |
| 2017 | Amara et al. [42] | *99.72* | *99.70* | *99.72* | *99.71* | 3 classes with 50/50 split |
| 2017 | Durmuú et al. [43] | 95.65 | - | - | - | 10 classes with 80/20 split |

Nisar Ahmed, Department of Computer Engineering, University of Engineering and Technology Lahore, Pakistan. E-Mail: nisarahmedrana@yahoo.com Contact: +92-300-7272402

| | | | | | | |
|---|---|---|---|---|---|---|
| 2017 | Brahimi et al. [44] | *99.185* | *98.529* | *98.532* | *98.518* | 9 classes with 5-fold CV |
| 2017 | Yamamoto et al. [45] | 92 | 90 | 92 | 91 | 9 classes with 80/20 split |
| 2018 | Yuan et al. [27] | 95.97 | - | - | - | 38 classes with 80/20 split |
| 2018 | Goncharov et al. [16] | *99* | - | - | - | 5 classes with 75/25 split |
| 2018 | Wang et al. [46] | 90.84 | - | - | - | 8 classes with 80/20 split |
| 2018 | Zhang et al. [28] | 91.79 | - | - | - | 38 classes with 80/20 split |
| 2018 | Ferentinos et al. [12] | **99.53** | - | - | - | 58 classes with 80/20 split |
| 2018 | Yadav et al. [47] | 97.39 | 97.76 | 97.39 | 97.34 | 23 classes with 70/30 split |
| 2019 | Kour et al. [48] | 95.23 | - | - | - | 7 classes with 70/15/15 split |
| 2018 | Pardede et al. [49] | 87.01 | - | - | - | 7 classes with 80/20 split |
| 2018 | Suryawati et al. [50] | 95.24 | - | - | - | 10 classes with 80/10/10 split |
| 2019 | Khandelwal et al. [13] | **99.37** | - | - | - | 57 classes with 80/20 split |
| 2019 | Barbedo et al. [11] | 94 | - | - | - | 14 classes with 80/20 split |
| 2019 | Baranwal et al. [51] | 98.54 | - | - | - | 38 classes with 80/20 split |
| | Proposed | 99.13±0.57 | 98.49±0.59 | 98.79 ±0.57 | 98.81 ±0.58 | 38 classes with 10-fold CV |

It can be noted from Table 13 that 7 existing works has claimed higher accuracy than the proposed scheme but four of them has used subset of the PlantVillage dataset and targeted a specific plant for disease identification. However, three of them have used similar or higher number of classes for disease identification. The three schemes with higher accuracies have used deep learning-based methods which employs image augmentation to increase the training dataset size. These methods take a longer time to train and require more computational and storage space to perform prediction which is a limitation for mobile devices as the ultimate utilization of the proposed scheme is deployment on mobile devices. Moreover, the testing scheme used in our case is 10-fold cross-validation which is a more rigorous method to check the behavior of classification performance as compared to hold-out validation such as 80/20 or 60/40. We have provided the standard deviation of values along with the performance metrics which are obtained over 10 train test splits. Moreover, the proposed algorithm work in a two-stage configuration and direct comparison of the proposed approach with other methods is not possible. There are 12 healthy leaf classes and 26 diseased leaf classes and we have performed the classification of diseased leaf only in the second stage. Further classification of the healthy leaf images is not in the scope of this work as we have targeted only to identify if a leaf image is healthy or diseased and then identify the category of disease if it is a diseased leaf image.

After finalizing the proposed model and evaluation of its results on PlantVillage dataset which is collected in a controlled environment, it is important to test available model on data collected independently for generalization. The images from online sources and some small dataset are gathered in self-collected dataset having same classes as in PlantVillage. Some images have been collected from the agricultural extension services and a set of 227 images is gathered. These images are selected based on their resemblance with the images in the database as more the variability in training and testing database, the more is the gap between their performances. The proposed model is tested on this dataset and a classification accuracy of 91.40% is achieved for healthy and unhealthy classification. In case of disease category classification, the accuracy is 82.47%. Note that decrease in classification accuracy of plant disease is due to difficulty in segmentation of the diseased leaf area. An

Nisar Ahmed, Department of Computer Engineering, University of Engineering and Technology Lahore, Pakistan. E-Mail: nisarahmedrana@yahoo.com Contact: +92-300-7272402 improvement in segmentation of diseased leaf area can result in even better disease identification accuracy and provide more reliable solution. Some of the schemes described in Table 13 have cross-validated their approach on an independently collected database and reported decrease in the classification accuracy. The proposed approach in comparison to deep learning-based approaches require low memory and less computation thus making it feasible for deployment on mobile devices.

## V. Conclusion

In this study, a model for identification of plant disease based on leaf image is proposed. The proposed model works in two stages, one categorizes the leaf image either as healthy or diseased. The second stage identifies the disease affecting the leaf in case of unhealthy leaf only. The first stage uses a bag-of-features approach and provided a recall of 99.48%. The second stage involves extraction of diseased leaf area and calculates a set of 29 texture and color features. Feature standardization is used to reduce the bias due to feature range and feature selection, and is performed to select the most discriminatory features. An ensemble of classifiers involving SVM, K-NN, Naïve Bayes, random forest and artificial neural networks was trained on the selected feature set and the classifier with best performance is selected. The SVM provided best performance with polynomial kernel of degree three and final model is constructed. The classification accuracy of final model for disease classification is 98.79% with a standard deviation of 0.57 over 10-fold cross-validation. The proposed model's ability to generalize is evaluated by collecting a dataset of 227 images and using the proposed model for evaluation. The classification accuracy of diseased and healthy plant identification is 91.40% and for disease category identification is 82.47%. The proposed model is a prototype system and it can be used to train on different disease classification scenarios. The only limitation is that the system will be able to classify only images which are visually distinguishable and provided in the training database.

### *A. Future Work*

a. An integrated solution containing plant species identification using leaf image in the first stage and plant disease identification in second stage can be formulated to improve the usability of the proposed solution.

b. Specialized models for high value crops or specific crops with more number of disease can be designed to identify the disease and categorize its severity based on fine-grained recognition.

Nisar Ahmed, Department of Computer Engineering, University of Engineering and Technology Lahore, Pakistan. E-Mail: nisarahmedrana@yahoo.com Contact: +92-300-7272402

Nisar Ahmed, Department of Computer Engineering, University of Engineering and Technology Lahore, Pakistan. E-Mail: nisarahmedrana@yahoo.com Contact: +92-300-7272402

## Appendix A

The details of the texture features are provided as follows:

I. Uniformity (angular second moment): It measures the orderliness or uniformity of the gray-level distribution of the diseased image region. Images with smaller number of gray-levels have higher uniformity and is calculated with the below formula:

$$Uniformity = \sum_{i=1}^{N_g}\sum_{j=1}^{N_g} p(i,j)^2 \tag{A.1}$$

II. Entropy: It measure the amount of randomness or disorder in the image. It is almost inversely related to uniformity and represent the amount of information in the image. An image with higher number of gray-levels will have higher entropy value and it is calculated with the below formula:

$$Entropy = -\sum_{i=1}^{N_g}\sum_{j=1}^{N_g} p(i,j)log[p(i,j)] \tag{A.2}$$

III. Contrast: It represent the amount of gray-level variation in the segmented diseased region. A high value of this parameter represent the presence of noise, edges or wrinkles in the image and is calculated with the below formula:

Nisar Ahmed, Department of Computer Engineering, University of Engineering and Technology Lahore, Pakistan. E-Mail: nisarahmedrana@yahoo.com Contact: +92-300-7272402

$$Contrast = \sum_{i=1}^{N_g} (i-j)^2 \left\{ \sum_{i=1}^{N_g} \sum_{j=1}^{N_g} p(i,j) \right\} \tag{A.3}$$

IV. Dissimilarity: Dissimilarity or inverse difference moment normalized is closely related to contrast and represent local gray-level variation in the segmented diseased region. It is almost inversely related to homogeneity and is calculated with the below formula:

$$Dissimilarity = \sum_{i=1}^{N_g} (i-j) \left\{ \sum_{i=1}^{N_g} \sum_{j=1}^{N_g} p(i,j) \right\} \tag{A.4}$$

V. Inverse Difference Moment (Homogeneity): It measure the homogeneity or smoothness of the gray-level distribution in the segmented diseased region. It has almost inverse relation with the contrast and have usually larger value when contrast is smaller. It is calculated with below formula:

$$Homogeneity = \sum_{i=1}^{N_g} \sum_{j=1}^{N_g} \frac{p(i,j)}{1+(i-j)^2} \tag{A.5}$$

VI. Inverse Difference (ID): It is a slightly different measure of the smoothness in gray-level distribution of segmented diseased region. It is calculated with the below formula:

$$ID = \sum_{i=1}^{N_g} \sum_{j=1}^{N_g} \frac{p(i,j)}{1+|i-j|} \tag{A.6}$$

VII. Correlation: It measures the linear dependency of gray-levels on the neighboring pixels of segmented diseased region. It provides almost similar information as of autocorrelation and is calculated with the below formula:

$$Correlation = \sum_{i=1}^{N_g} \sum_{j=1}^{N_g} \frac{ijp(i,j) - \mu_x \mu_y}{\sigma_x \sigma_y} \tag{A.7}$$

VIII. Autocorrelation: It measure the correlation of the gray-levels with shifted version of the segmented diseased region and is closely related to correlation. It is computed with the following formula:

$$Autocorrelation = \sum_{i=1}^{N_g} \sum_{j=1}^{N_g} p(i,j)\bar{p}(i-k, j-k) \tag{A.8}$$

Where: $k$ is the amount of shift.

Nisar Ahmed, Department of Computer Engineering, University of Engineering and Technology Lahore, Pakistan. E-Mail: nisarahmedrana@yahoo.com Contact: +92-300-7272402

IX. Cluster Shade (CS): Cluster shade use third order moment to represent texture and show higher range as compared to other measures. For a homogeneous segmented diseased image region, cluster shade values are consistent and use compact range. It is computed with the following formula:

$$CS = \sum_{i=1}^{N_g}\sum_{j=1}^{N_g} p\left(i + j - \mu_x - \mu_y\right)^3 \bar{p}(i,j) \tag{A.9}$$

X. Cluster Prominence (CP): Cluster prominence use forth order moment to represent texture and have similar characteristics as cluster shade. It is computed with the following formula:

$$\text{CP} = \sum_{i=1}^{N_g}\sum_{j=1}^{N_g} p\left(i + j - \mu_x - \mu_y\right)^4 \bar{p}(i,j) \tag{A.10}$$

XI. Maximum Probability (MP): This measure is based on homogeneity and calculates the maximum probability of occurrence for all$(I,j)$. It is calculated with below formula:

$$MP = \max_{i,j} p(i,j) \tag{A.11}$$

XII. Sum of Squares (SS): It is a closely related feature to uniformity and is calculated with the below formula:

$$SS = \sum_{i=1}^{N_g}\sum_{j=1}^{N_g} p(i-\mu)^2 p(i,j) \tag{A.12}$$

XIII. Sum Average (SA): It measures the mean of the gray-level sum distribution of the segmented diseased region. It is a related measure of distribution of the sum of gray-level co-occurring pixels in the image and is calculated with the below formula :

$$SA = \sum_{i=2}^{2N_g} i p_{x+y}(i) \tag{A.13}$$

XIV. Sum Variance (SV): It measures the dispersion of gray-level sum distribution with respect to mean value of the segmented diseased region and has close relation to the distribution of sum of the co-occurring pixels. It is calculated with the below formula:

$$SV = \sum_{i=2}^{2N_g}\left(i - \left[\sum_{i=2}^{2N_g} p_{x+y}(i)\right]\right)^2 \tag{A.14}$$

Nisar Ahmed, Department of Computer Engineering, University of Engineering and Technology Lahore, Pakistan. E-Mail: nisarahmedrana@yahoo.com Contact: +92-300-7272402

XV. Sum Entropy (SE): It measure the amount of randomness or disorder in the gray-level sum distribution of segmented diseased region. It is a measure related to the distribution of the sum of co-occurring pixels in the image and is calculated with the below formula.

$$SE = -\sum_{i=2}^{2N_g} p_{x+y}(i) log\{p_{x+y}(i)\} \tag{A.15}$$

XVI. Difference Variance (DV): It measure the amount of dispersion with respect to mean of the gray-level difference distribution of segmented diseased region. It is a measure related to the distribution of the difference between co-occurring pixels in the image and is calculated with the below formula.

$$\boldsymbol{DV = \sum_{i=2}^{2N_g} \left(i - \left[\sum_{i=2}^{2N_g} ip_{x-y}(i)\right]\right)^2} \tag{A.16}$$

XVII. Difference Entropy (DE): It measure the amount of randomness or disorder in relation to the gray-level difference distribution of segmented diseased region. It is a measure related to the distribution of the difference between co-occurring pixels in the image and is calculated with the below formula.

$$\boldsymbol{DE = -\sum_{i=0}^{N_g-1} p_{x-y}(i) log\{p_{x-y}(i)\}} \tag{A.17}$$

XVIII. Information Measures of Correlation (IMC): These measures of correlation have some desirable properties which are represented by rectangular correlation measure. There are two information measures of correlation and are calculated with below formulas:

$$\boldsymbol{IMC_1 = \frac{HXY - HXY_1}{max\{HX, HY\}}} \tag{A.18}$$

$$\boldsymbol{IMC_2 = \sqrt{1 - e^{[-2.0(HXY_2 - HXY)]}}} \tag{A.19}$$

Where; $HXY = \sum_{i=1}^{Ng} \sum_{j=1}^{Ng} p(i,j) \log(p(i,j))$

XIX. Maximal Correlation Coefficient (MCC): It is a measure of dependence of two random variables and it defines the least upper bound of the values of the correlation coefficients. It is calculated with the below formula:

$$\boldsymbol{MCC = \sqrt{F\left(\sum_k \frac{p(i,k)p(j,k)}{p_x(i)p_y(k)}\right)}} \tag{A.20}$$

Where; $F$ is the second largest Eigen value.

Nisar Ahmed, Department of Computer Engineering, University of Engineering and Technology Lahore, Pakistan. E-Mail: nisarahmedrana@yahoo.com Contact: +92-300-7272402

XX. Inverse Difference Normalized (IDN): It provides normalized value of homogeneity and the normalization is done with the number of gray-levels which is eight in our case. It is calculated with the below formula:

$$IDN = \sum_{i,j=1}^{G} \frac{C_{ij}}{1 + |i-j|^2 / G^2} \qquad \text{(A.21)}$$

XXI. Inverse Difference Moment Normalized (IDMN): It provides normalized second moment of the homogeneity and the value of IDMN and IDN is closely related to contrast and dissimilarity measures. It is calculated with the below formula:

$$IDMN = \sum_{i,j=1}^{G} \frac{C_{ij}}{1 + (i-j)^2 / G^2} \qquad \text{(A.22)}$$

## APPENDIX B

*Evaluation Measure*

Confusion Matrix: It is a matrix which describes the classifier performance on testing data and is also known as error matrix. It is called confusion matrix as it allows to spot the classes which may be confused by the classifier. Each column of the matrix represents the examples in actual class and each row represent the examples in predicted class.

Precision & Recall: In binary classification, precision is the portion of relevant examples among the retrieved examples whereas recall is the portion of relevant examples which are retrieved over the total number of relevant instances. Both of these parameters are based on the measure of relevance and are essential to describe a classifier’s performance. Formulas for calculation of precision and recall are provided below:

$$Precision = \frac{True\ Possitive}{True\ Positive + False\ Positive} \qquad \text{(B.1)}$$

$$Recall = \frac{True\ Possitive}{True\ Positive + False\ Negative} \qquad \text{(B.2)}$$

Accuracy: It is most commonly used measure for evaluation of classifier performance. It measures the ratio of correctly classified observations to the total observations. The formula for accuracy computation is provided below:

$$Accuracy = \frac{True\ Possitive + True\ Negative}{Total\ Population} \qquad \text{(B.3)}$$

F1-Score: It measures the harmonic mean between precision and recall and sometimes considered a better measure of performance. It is a useful measure to check classifier performance than accuracy especially when class distribution is uneven as it takes both false positive and false negatives into account. The formula for F1-score computation is provided below:

$$\boldsymbol{F1-Score = 2 \times \frac{Precision \times Recall}{Precision + Recall}} \quad \text{(B.4)}$$

Nisar Ahmed, Department of Computer Engineering, University of Engineering and Technology Lahore, Pakistan. E-Mail: nisarahmedrana@yahoo.com Contact: +92-300-7272402